\documentclass[11pt,a4paper]{article}
\usepackage[hyperref]{emnlp2020}
\usepackage{times}
\usepackage{latexsym}

\usepackage{microtype}

\usepackage{graphicx} %SP
\usepackage{subfig}%JS
\usepackage{color} %SP
\usepackage{comment} % SP
\usepackage{gensymb} %RF
\usepackage{booktabs} %RF
\usepackage{colortbl} %RF
\usepackage{enumitem} %RF
\usepackage{arydshln} % RF

\usepackage{siunitx} %MG
\usepackage{multirow} %ET

\aclfinalcopy % Uncomment this line for the final submission
\def\aclpaperid{2058} %  Enter the acl Paper ID here

\title{\emph{Refer, Reuse, Reduce}\\ Generating Subsequent References in Visual and  Conversational Contexts}

\author{Ece Takmaz, Mario Giulianelli, Sandro Pezzelle, Arabella Sinclair, Raquel Fern\'andez \\
  Institute for Logic, Language and Computation \\
  University of Amsterdam \\
  \texttt{\{e.takmaz|m.giulianelli|s.pezzelle|}\\
  \texttt{a.j.sinclair|raquel.fernandez\}@uva.nl}}

\date{}

\begin{document}
\maketitle
\begin{abstract}
Dialogue participants often refer to entities or situations repeatedly within a conversation, which contributes to its cohesiveness. Subsequent references exploit the common ground accumulated by the interlocutors and hence have several interesting properties, namely, they tend to be shorter and reuse expressions that were effective in previous mentions. 
In this paper, we tackle the generation of first and subsequent references in visually grounded dialogue. We propose a generation model that produces referring utterances grounded in both the visual and the conversational context. To assess the referring effectiveness of its output, we also implement a reference resolution system. Our experiments and analyses show that the model produces better, more effective referring utterances than a model not grounded in the dialogue context, and generates subsequent references that exhibit linguistic patterns akin to humans.
\end{abstract}
%================================

% !TEX root = main_pbgen.tex

%==================================
\section{Introduction}
\label{sec:intro}
%==================================

\begin{figure}[t]\centering \small
\includegraphics[height=3.2cm]{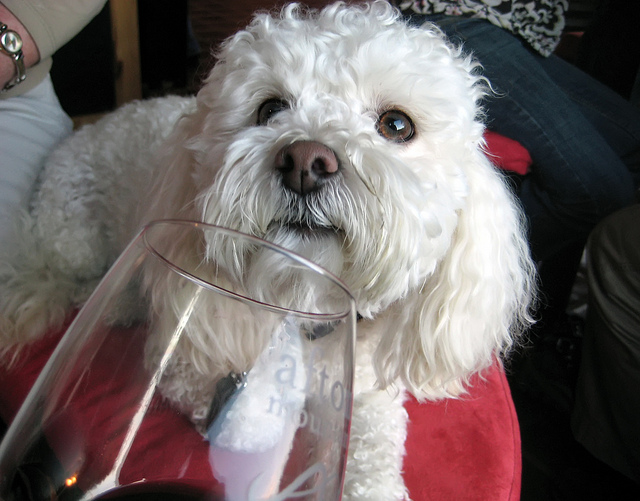}\\
\vspace*{2pt}
\fbox{\begin{minipage}{.95\columnwidth}
\begin{tabular}{@{}l@{\ }l}
& \textbf{Referring utterances extracted from dialogue 1}\\[2pt]
                  & A: a white fuzzy dog with a wine glass up to his face\\
$\leadsto$ & B: I see the wine glass dog\\
$\leadsto$ & A: no I don't have the wine glass dog
\end{tabular}
\end{minipage}
}\\[1pt]
\fbox{\begin{minipage}{.95\columnwidth}
\begin{tabular}{@{}l@{\ }l}
& \textbf{Referring utterances extracted from dialogue 2}\\[1pt]
                  & C: white dog sitting on something red\\
$\leadsto$ & D: yes I have the dog on the red chair\\
$\leadsto$ & C: white dog  on the red chair
\end{tabular}
\end{minipage}
}
\caption{Two chains of referring utterances from two games with different participants, including the first description of the target image in that dialogue and two subsequent references ($\leadsto$). In the game, each participant sees 5 additional images besides the target shown here. The distractor images change at every round of the game, i.e., each co-referring utterance within a dialogue is produced in a slightly different visual context.\label{fig:ex1}}
\end{figure}

When speakers engage in conversation, they often refer to the same objects or situations more than once. Subsequent references  \cite{mcdonald-1978-subsequent-reference} are dependent on the shared knowledge that speakers accumulate during dialogue. For example, dialogue participants may first mention \emph{``a white fuzzy dog with a wine glass up to his face''}  and later refer to it as  \emph{``the wine glass dog''}, as shown in  Figure~\ref{fig:ex1}, dialogue 1. Speakers establish `conceptual pacts', i.e., particular ways of conceptualising referents that condition what is perceived as coherent in a given dialogue \cite{GarrodAnderson1987,BrennanClark1996}. While \emph{``the wine glass dog''} may be odd as a standalone description, it is an appropriate referring expression in the above conversational context. Yet,
uttering it in a different context (such as dialogue 2 in Figure~\ref{fig:ex1}, after the participants had successfully referred to the image as \emph{``the dog on the red chair''})
may 
disrupt the cohesion of the dialogue and lead to communication problems \cite{metzing2003conceptual}.

%------------ FIGURE -----------------
%-------------END OF FIG---------------

In this paper, we tackle the generation of referring utterances---i.e., utterances that contain referring descriptions, as in Figure~\ref{fig:ex1}---grounded both in the visual environment and the dialogue context. 
 These utterances have several interesting properties that make their automatic generation challenging. First, they are produced with the communicative goal of helping the addressee identify the intended referent.
 Second, because humans operate under cognitive and time-bound constraints,
 dialogue participants will aim to fulfil this communicative goal while optimising the use of their limited cognitive resources. This results in two common features of subsequent mentions: 
 (1)~\emph{Reduction:} Utterances tend to become shorter---a well attested phenomenon since the 
 work of \citet{KraussWeinheimer1967}---as a result of interlocutors' 
 reliance on their common ground \cite{stalnaker2002common}: As more shared information is accumulated, it becomes predictable and can 
 be left implicit \cite{Grice75,ClarkWilkes-Gibbs1986,clark1991grounding,Clark1996}. Sentence compression also takes place in discourse, as predicted by the entropy rate principle \cite{genzel-charniak-2002-entropy,keller-2004-entropy}. 
 (2)~\emph{Lexical entrainment:} Speakers tend to reuse words that were effective in previous mentions \cite{GarrodAnderson1987,BrennanClark1996} possibly due to priming effects \cite{Pickering2004-PICTAM}. 
Thus, besides being a challenging problem intriguing from a linguistic and psycholinguistic point of view, computationally modelling the generation of subsequent references can contribute to better user adaptation in dialogue systems and to more natural human-computer interaction.

For our study, we use data from the PhotoBook dataset \cite{haber2019photobook}, 
developed to elicit subsequent references to the same images within task-oriented dialogue. 
To isolate the issue we are interested in, 
we extract, from each dialogue, the utterances that refer to a given image. This results in a dataset of dialogue-specific chains of co-referring utterances: For example, Figure~\ref{fig:ex1} shows two chains of co-referring utterances from two different dialogues, both referring to the same image. Figure~\ref{fig:extraction} shows another example.
We then formulate the problem as 
the generation of the next utterance in a chain given the current visual context and the common ground established in previous co-referring utterances  (whenever these are available). 
To computationally model this problem, we propose three variants of a generation system based on the encoder-decoder architecture \cite{Sutskever}.  
We evaluate their output with metrics commonly used in the domain of Natural Language Generation and with several linguistic measures. In addition, to assess the communicative effectiveness of the generated references, we implement a reference resolution agent in the role of addressee.

We find that conditioning the generation of referring utterances on previous mentions leads to better, more effective descriptions than those generated by a model that does not exploit the conversational history.
Furthermore, our quantitative and qualitative analysis shows that the context-aware model generates subsequent references that exhibit linguistic patterns akin to humans' regarding markers of new vs.\ given information, reduction, and lexical entrainment, including novel noun-noun compounds.  

Our data, code, and models are available at \url{https://dmg-photobook.github.io}.

% !TEX root = main_pbgen.tex

%==================================
\section{Related Work}
\label{sec:relatedwork}
%==================================

%----------------------------------------------------------------------
\paragraph{Generation of distinguishing expressions}
%----------------------------------------------------------------------

Our work is related to Referring Expression Generation (REG), a task with a long tradition in computational linguistics that consists in generating a description that distinguishes a target from a set of distractors---\citet{krahmer-van-deemter-2012-computational} provide an overview of early approaches. 
Follow-up approaches focused on more data-driven algorithms exploiting datasets of simple visual scenes annotated with symbolic attributes~\cite[e.g.,][among others]{mitchell2013typicality,mitchell2013generating}.  
More recently, the release of large-scale datasets with real images~\cite{kazemzadeh2014referitgame} has made it possible to test deep learning multimodal models on REG,  sometimes in combination with referring expression comprehension~\cite{mao2016generation,Yu2017AJS}.
While REG typically focuses on describing objects within a scene, 
a few approaches at the intersection of REG and image captioning~\cite{bernardi2016automatic} have aimed to generate discriminative descriptions of full images, i.e., image captions that can distinguish the target image from a pool of related ones~\cite{andreas-klein-2016-reasoning,vedantam2017context,cohn-gordon-etal-2018-pragmatically}.
Similarly to these approaches, in the present work, we
generate utterances that refer to a full image with the aim of  distinguishing it from other distractor images. In addition, our setup has several novel aspects: The referring utterances are the result of interactive dialogue between two participants and include 
subsequent references.

%----------------------------------------------------------------------
\paragraph{Generation of subsequent references}
%----------------------------------------------------------------------
Follow-up work within the REG tradition has extended the early algorithms to deal with 
subsequent references~\cite{gupta2005automatic,jordan2005learning,stoia-etal-2006-noun,viethen-etal-2011-generating}. 
These approaches focus on content selection (i.e., on generating a list of attribute types such as \texttt{color} or \texttt{kind} using an annotated corpus) or on choosing the type of reference (definite or indefinite noun phrase, pronoun, etc.) 
and do not directly exploit visual representations. In contrast, we generate the surface realisation of first and subsequent referring utterances end-to-end, grounding them in continuous visual features of real images.

Our work is related to a recent line of research on reference \emph{resolution} in visually-grounded dialogue, where previous mentions have been shown to be useful~\cite{shore-skantze-2018-using,haber2019photobook,roy-etal-2019-leveraging}. 
Here we focus on \emph{generation}. To our knowledge, this 
is the first attempt at generating visually grounded referring utterances taking into account earlier mentions in the dialogue.  
Some work on generation 
has exploited dialogue history in order to make lexical choice decisions that align with what was said before \cite{brockmann2005modelling,buschmeier-etal-2009-alignment,stoyanchev-stent-2009-lexical,lopes2015rule,hu2016entrainment,dusek-jurcicek-2016-context}. Indeed, incorporating entrainment in dialogue systems leads to an increase in the perceived naturalness of the system responses and to higher task success \cite{lopes2015rule,hu2016entrainment}.
As we shall see, our generation model exhibits some lexical entrainment.

\paragraph{Dialogue history in visual dialogue}
%----------------------------------------------------------------------

Recent work in the domain of visually grounded dialogue has exploited dialogue history in encoder-decoder models trained on large datasets of question-answering dialogues \cite{das2017visual,de2017guesswhat,chattopadhyay2017evaluating}.
Recently, 
\citet{agarwal2020history} showed that only 10\% of the questions in the VisDial dataset \cite{das2017visual} genuinely require dialogue history in order to be answered correctly, which is in line with other shortcomings highlighted by~\citet{massiceti2018}. More generally, visually grounded dialogue datasets made up of sequences of questions and answers lack many of the collaborative aspects that are found in natural dialogue. For our study, we focus on the PhotoBook dataset by \citet{haber2019photobook}, where dialogues are less restricted and where the common ground accumulated over the dialogue history plays an important role.

% !TEX root = main_pbgen.tex

%==================================
\section{Data}
\label{sec:data}
%==================================

%------------------------------------------
\subsection{PhotoBook Dataset}
\label{sec:pb}
%------------------------------------------

The PhotoBook dataset \citep{haber2019photobook} is a collection of task-oriented visually grounded English dialogues between two participants. The task is set up as a game comprised of 5 rounds. In each round, the two players are assigned private `photo books' of 6 images, with some of those images being present in both photo books. The goal 
is to find out which images are common to both players by interacting freely using a chat interface. In each round, the set of 6 images available to each player changes, but a subset of images reappears, thus triggering subsequent references to previously described images. 
This feature of the PhotoBook dataset makes it a valuable resource to model the development of conversational common ground between interlocutors.
The dataset consists of 2,500 games, 165K utterances in total, and 360 unique images 
from MS COCO~\cite{lin2014coco}. 

%------------------------------------------
\subsection{Dataset of Referring Utterance Chains}
\label{sec:extraction}
%------------------------------------------

\begin{figure*}\small \centering
\fbox{ 
\begin{minipage}{5.5cm}
\includegraphics[width=2.5cm]{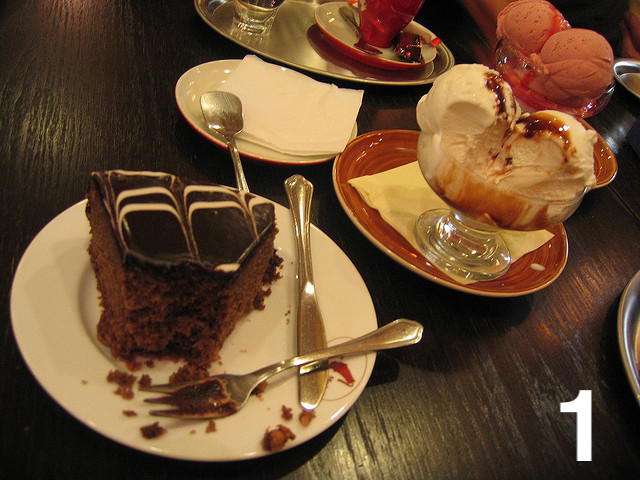} \includegraphics[width=2.5cm]{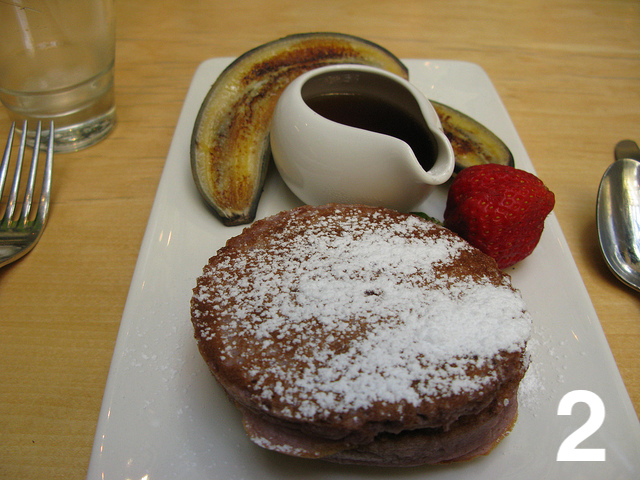}\\
\includegraphics[width=2.5cm]{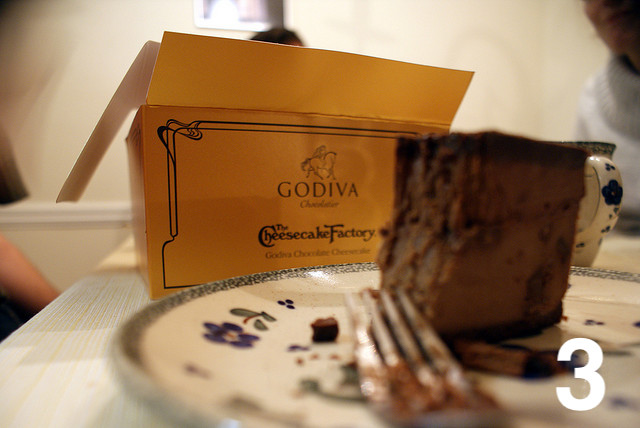}
\includegraphics[width=2.5cm]{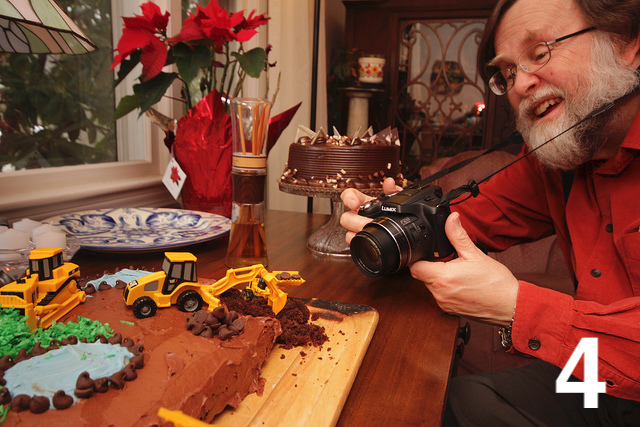}\\
\includegraphics[width=2.5cm]{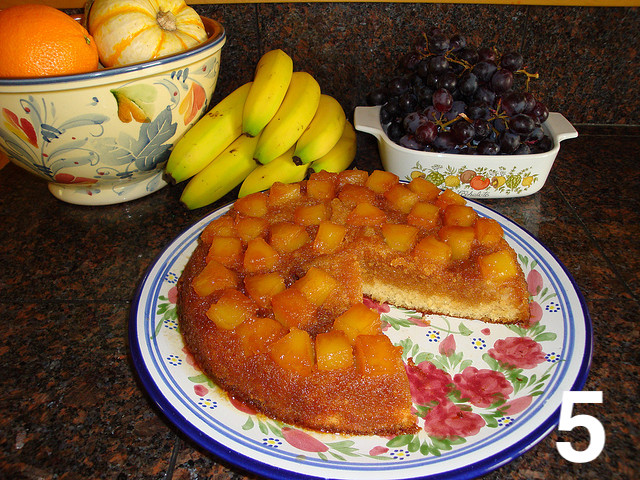}
\includegraphics[width=2.5cm]{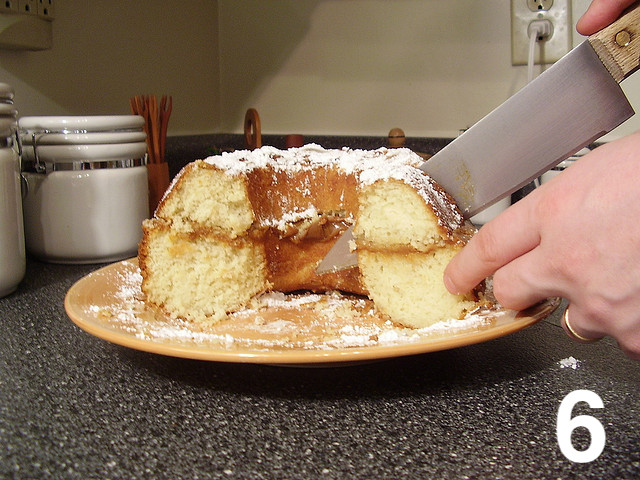}
\end{minipage}
\begin{minipage}{9.5cm}
\begin{minipage}{9cm}
\textsc{Dialogue fragment and images visible to participant A}\\
\textsc{in the first round of a game}\\[5pt]
\begin{tabular}{l@{\ }l}
A: & Hi\\
B: & Hello.\\
B: & do you have a white cake on multi colored striped cloth?\\
A: & \textbf{I see a guy taking a picture. What about you?}\\
B: & is it of a cake with construction trucks on it?\\
A: & Yeah. I don't see the cake you mentioned.\\
A: & \texttt{<common img\_4>}
\end{tabular}
\end{minipage}\\[5pt]
\begin{minipage}{9cm}
\textsc{Resulting referring utterance chain with subsequent}\\
\textsc{references extracted from the following game rounds}\\[5pt]
\begin{tabular}{l}
1. I see a guy taking a picture. What about you?\\
2.  guy with camera\\
3.  I have the guy with camera\\
4.  The last one is the camera guy.
\end{tabular}
\end{minipage}
\end{minipage}
}
\caption{
Example from our new dataset of referring utterance chains. Given a target image selected by a participant (here \texttt{<common img\_4>}), the utterances in the dialogue prior to that selection action are scored by their likelihood of referring to the target. In this example, the utterance in bold is selected as the first description. To construct the reference chain, subsequent references are extracted in a similar manner from the dialogue in the following game rounds. The set of distractor images available to a participant changes across rounds.
\label{fig:extraction}}
\end{figure*}

As mentioned above, in PhotoBook participants can freely interact via chat. The dialogues thus include different types of dialogue act besides referring utterances. 
While utterances performing other functions are key to the dialogue and may provide useful information, in the present work we abstract away from this aspect and concentrate on referring utterances.\footnote{\citet{haber2019photobook}  extracted co-reference chains made up of multi-utterance dialogue excerpts. Our  chains include single utterances, which is more suitable for generation.}
To create the data for our generation task, we extract utterances that contain an image description and their corresponding image target from the dialogues as follows.  Within a game round, we consider all the utterances up to the point where a given image $i$ has been identified by the participants\footnote{Image identification actions are part of the metadata.}
as candidate referring utterances for \emph{i} -- see Figure~\ref{fig:extraction}. 
We then compare each candidate against a reference set of descriptions made up of the MS COCO \citep{lin2014coco} captions for \emph{i} and the attributes and relationship tokens of \emph{i} in the Visual Genome \citep{krishna2017visual}. We score each candidate utterance  with the sum of its BERTScore\footnote{BERTScore uses contextualised embeddings \cite{devlin-etal-2019-bert} to assess similarity between a target sentence and one or more reference sentences.} \citep{bert-score}
 for captions and its METEOR score \citep{banerjee2005meteor} for attributes and relationships. The top-scoring utterance in the game round is selected as a referring utterance for \emph{i} and used as an additional caption for extracting subsequent references in the following game rounds. 
As a result of this procedure, for a given dialogue and an image \emph{i}, we obtain a reference chain  made up of the referring utterances---maximum one per round---that refer to \emph{i} in the dialogue. 
Since images do not always reappear in each round, chains can have different length. Two examples of chains of length 3 are shown in Figure~\ref{fig:ex1} and a chain of length 4 in Figure~\ref{fig:extraction}. Given that each utterance in a chain belongs to a different game round, each utterance was produced in a slightly different visual context with different distractor images. Figure~\ref{fig:extraction} shows the visual context available to participant A in the first round of a game, when the participant produced the first description in the dialogue for target image number 4. The other three descriptions in the chain were produced while seeing different distractors. 

We evaluate the referring utterance extraction procedure and the resulting chains using 20 dialogues hand-annotated by \citet{haber2019photobook} with labels linking utterances to the target image they describe. Using our best setup, we obtain a precision of 0.86 and a recall of 0.61. The extracted chains are very similar to the human-annotated ones in terms of chain and utterance length. 

Our new dataset is made up of 41,340 referring utterances and 16,525 chains (i.e., there are 16,525 first descriptions and 24,815 subsequent references). The median number of utterances in a chain is 3.
We use the splits defined by \citet{haber2019photobook} to divide the dataset into Train, Validation, and Test, and  all hand-annotated dialogues are excluded from these splits. Table \ref{tab:splits} reports relevant descriptive statistics of the dataset. 
More details about the extraction procedure and the dataset are available in Appendix \ref{app:data}.
Appendix~\ref{sec:proc} describes how the dataset is further processed to be used in our models.

\begin{table}\centering 
	\resizebox{0.95\columnwidth}{!}{%
		\begin{tabular}{| l | r | r | c | r | c |}
			\hline
			\multicolumn{1}{|c|}{\multirow{2}{*}{\bf Split}}  & \multirow{2}{*}{\bf Games}  &  \multicolumn{2}{c|}{\bf \emph{First}} & \multicolumn{2}{c|}{\bf \textit{Later}} \\ \cline{3-6}
			&  & \multicolumn{1}{c|}{\bf N} & \bf Length & \multicolumn{1}{c|}{\bf N} & \bf Length \\  \hline
			\bf Train & 1725 & 11540 & 10.52 (4.80) & 17393 & 7.52 (4.15) \\ \hline
			\bf Val    & 373 & 2503 & 10.49 (4.81) & 3749 & 7.70 (4.22) \\ \hline
			\bf Test  & 368 & 2482 & 10.52 (4.85) & 3673 & 7.59 (4.17) \\ \hline
		\end{tabular}
	}
	\caption{Number of games and referring utterances in the splits of our dataset with their average length in tokens (standard deviation in brackets) broken down by first mentions vs.~subsequent (`Later') references.}
	\label{tab:splits}
\end{table}

% !TEX root = main_pbgen.tex

%==================================
\section{Models}
\label{sec:model}
%==================================

With the new dataset of referring utterance chains in place, we operationalise the problem of generating a referring utterance taking into account the visual and conversational context as follows.
The model aims to generate a referring utterance given (a) the \emph{visual context} in the current game round made up of $6$ images from the perspective of the player who produced the utterance, (b) the \emph{target} among those images, and (c) the \emph{previous co-referring utterances} in the chain (if any). 
Besides being contextually appropriate, the generated utterance has to be informative and discriminative enough to allow an addressee to identify the target image. 
We thus also develop a reference resolution model that plays the role of addressee. 
The two models are trained independently.

%==================================
\subsection{Generation Models}
%==================================

We propose three versions of the generation model, which all follow the encoder-decoder architecture~\cite{Sutskever}. These versions differ from
each other with respect to whether
and how they exploit earlier referring utterances for the target image: (1) a baseline model that does not use the dialogue context at all (hence, \textbf{Ref}); (2) a model that conditions the generation on the  previous referring utterance, if available, and operates attention over it (hence, \textbf{ReRef}); (3) a model that builds on (2) by adding a `copy' mechanism~\cite{GTTP} (hence, \textbf{Copy}). 
We describe them below and provide further details in Appendix \ref{sec:models}.

%----------------------
\paragraph{Ref}
%----------------------

This model is provided only with the information about the visual context in the current game round--and not with the linguistic context in previous rounds. We encode each image in the context by means of visual features extracted from the penultimate layer of ResNet-152~\cite{7780459} pretrained on ImageNet~\cite{imagenet_cvpr09}. First, the visual features of the $6$ candidate images are concatenated. This concatenated vector goes through dropout, a linear layer and ReLU~\cite{relu}. The same process is applied for the single target image.
We then concatenate the final visual context vector with the target image vector, 
apply a linear transformation, and use the resulting hidden representation $h_d$ 
to initialise an LSTM decoder, which generates the referring utterance one word at a time. At each timestep, the input to the decoder is a multimodal vector, i.e., the concatenation of $h_d$ and the word embedding of token $t_t$.
The weights of the embeddings are initialised uniformly in the range $(-0.1, 0.1)$ and learned from scratch for the task at hand.

%------------------------------------------------------------
\paragraph{ReRef}
%------------------------------------------------------------

With this model, we aim to simulate a speaker who is able to \emph{re-refer} to a target image in accordance with what has been established in the conversational common ground \cite{Clark1996,BrennanClark1996}. The model enriches {\bf Ref} by incorporating linguistic information into the encoder (in addition to visual information) and an attention mechanism applied over the hidden states of the encoder during decoding. The model thus generates a new utterance conditioned on both the visual and the linguistic context.

The encoder is a one-layer bidirectional LSTM initialised with the same visual input fed to {\bf Ref}. In addition, it receives as input the previous referring utterance used in the dialogue to refer to the target image,\footnote{The latest description seems to contain the most relevant information. Including all referring utterances in the chain up to that point in the dialogue did not lead to improvements.}
or else is fed the special $<$nohs$>$ token, indicating that there is no conversational history for the target image yet. 
We utilise the attention mechanism proposed by \citet{Bahdanau} and used by \citet{GTTP}. 
During decoding, attention contributes to determining which aspects of the multimodal context are most critical in generating  the next referring utterance. We expect this attention mechanism to be able to identify the words in a previous utterance that should be present in a subsequent reference, resulting in lexical entrainment.

%----------------------
\paragraph{Copy}
%----------------------

This model builds on {\bf ReRef} and incorporates a means of simulating 
lexical entrainment more explicitly, by regulating when a word  used in the previous mention should be used again in the current referring utterance (i.e., should be produced by the decoder). 
Given the shortening property of subsequent references mentioned in the Introduction, our task bears some similarity to text summarisation. We thus draw inspiration from the summarisation model proposed by~\citet{GTTP}. 
In particular, we equip the model with their `copy' mechanism, 
which combines the probability of copying a word present in the encoded input 
with the probability of generating that word from the vocabulary. We expect this mechanism to contribute to generating rare 
words present in preceding referring utterances that are part of a `conceptual pact' \cite{BrennanClark1996} between the dialogue participants, but may have low generation probability overall.

%==================================
\subsection{Reference Resolution Model}
%==================================

Given an utterance referring to a target image and a $6$-image visual context, our reference resolution model
predicts the target image among the candidates. This model is similar to the resolution model proposed by~\citet{haber2019photobook} for the PhotoBook dataset, but includes several extensions:
(1)~We use BERT embeddings from the uncased base BERT model~\cite{devlin-etal-2019-bert, Wolf2019HuggingFacesTS} to represent the linguistic input rather than LSTMs;\footnote{In the generation models, we did not use BERT due to the difficulties of using contextualised embeddings in the decoder, and the desirability of 
using the same word embeddings in both the encoder and the decoder.}
(2)~The input utterance is encoded taking into account the visual context: We compute a multimodal representation of the utterance by concatenating each BERT token representation with the visual context representation, obtained in the same way as for the generation models;\footnote{We also tried using multimodal representations obtained via LXMERT~\cite{tan-bansal-2019-lxmert}. 
No improvements were observed.}
(3)~We apply attention over the multimodal representations of the utterance in the encoder 
instead of using the output from a language-only LSTM encoder. The utterance's final representation is given by the weighted average of these multimodal representations with respect to the attention weights.

Each candidate image 
is represented by its ResNet-152 features~\cite{7780459} or, if it has been previously referred to in the dialogue, by the sum of the visual features and the representation of the previous utterance (obtained via averaging its BERT embeddings).\footnote{Thus, some of the candidate images have multimodal representations (if they were already mentioned in the dialogue), while others do not.} 
To pick a referent, we take the dot product between the representation of the  input utterance 
and each of the candidate image representations. The image with the highest dot-product value is the one chosen by the model.

%==================================
\subsection{Model Configurations}
%==================================

For each model, we performed hyperparameter search for batch size, learning rate, and dropout; also, the search included different dimensions for the embedding, attention, and hidden layers. All models were trained for up to $100$ epochs 
(with a patience of $50$ epochs in the case of no improvement to the validation performance) using the Adam optimiser~\cite{Adam} to minimise the Cross Entropy Loss with sum reduction. 
BERTScore F1~\cite{bert-score} in the validation set was used to select the best model for the generation task, while we used accuracy for the resolution task. 
In the next section, we report average scores and standard deviations over $5$ runs with different random seeds. 
Further details on hyperparameter selection, model configurations, and reproducibility can be found in Appendix \ref{sec:config}.

% !TEX root = main_pbgen.tex

%==================================
\section{Results}
\label{sec:results}
%==================================

%---------------------------------------------
\subsection{Evaluation Measures}
%---------------------------------------------

We evaluate the performance of the reference resolution model by means of both accuracy and Mean Reciprocal Rank (MRR). As for the generation models, we compute several metrics that are commonly used in the domain of Natural Language Generation. In particular, we consider three measures based on \emph{n-}gram matching:  BLEU-2~\cite{Papineni:2002},\footnote{BLEU-2, which is based on bigrams, appears to be more informative than BLEU with longer $n$-grams in dialogue response generation \cite{liu-etal-2016-evaluate}}
ROUGE~\cite{Lin2004}, and CIDEr~\cite{cider}. We also compute BERTScore F1~\cite{bert-score} (used for model selection), which in our setup 
compares the contextual embeddings of the generated sentence to those of the set of referring utterances in the given \emph{chain}. Further details of the metrics are in Appendix \ref{app_evalmetrics}.

All these measures capture the degree of similarity between generated referring utterances and their human counterparts. 
In addition, to assess the extent to which the generated utterances fulfil their communicative goal, we pass them to 
our reference resolution model and obtain accuracy and MRR. While this is not a substitute for human evaluation, we take it to be an informative proxy. In Section~\ref{sec:analysis}, we analyse the generated utterances with respect to linguistic properties related to phenomena that are not captured by any of these metrics.

%------------------------------------
\subsection{Reference Resolution Results}
%------------------------------------

Our reference resolution model achieves an 
accuracy of 85.32\% and MRR of 91.20\% on average over 5 runs. This is a substantial result. 
A model that predicts targets at random would yield an accuracy of roughly 
16.67\% (as the task is to pick one image out of 6 candidates), while a baseline that simply takes one-hot representations of the image IDs in the context achieves 
22.37\% accuracy.\footnote{In this simple baseline, one-hot vectors are projected to scalar values, and a softmax layer assigns probabilities over them. The fact that this is slightly higher than random accuracy seems due the different frequencies of images being the target.} 

\begin{table}[h!]\centering 
	\resizebox{0.9\columnwidth}{!}{%
		\begin{tabular}{| l | r | r | c|}\hline
			\bf Subset & \multicolumn{1}{c|}{\bf ACC} & \multicolumn{1}{c|}{\bf MRR} & \multicolumn{1}{c|}{\bf Instances}\\ \hline
			\bf First  & 80.27 (0.46) & 87.78 (0.28) & 2482\\\hline
			\bf Later  & 88.74 (0.18) & 93.51 (0.09)& 3673\\\hline
			\bf Overall & 85.32 (0.19) &91.20 (0.10) &6155\\\hline
		\end{tabular}%
	}
	\caption{Test set scores of the reference resolution model: averages of 5 runs with the best configuration, with the standard deviations in parentheses.}
	\label{tab:resolution}
\end{table}	

In Table ~\ref{tab:resolution}, the results are presented by breaking down the test set into two subsets: the \emph{first} referring utterances in a chain, and \emph{later} referring utterances, i.e., subsequent references where the target image among the candidates has linguistic history associated with it. The model performs better on subsequent references. Exploiting dialogue history plays a role in this boost: an ablated version of the model that does not have access to the linguistic history of subsequent references yields an accuracy of 84.82\% for the \emph{Later} subset, which is significantly lower than the 88.74\% obtained with our model ($p< 0.01$ independent samples $t$-test). This confirms the importance of accessing information about previous mentions in visually grounded reference resolution~\cite{haber2019photobook}. 

We use the best model run to assess the communicative effectiveness of our generation models.

\begin{table*}[h!]
	\resizebox{\textwidth}{!}{%
		\begin{tabular}{|l|l|c|c|c|c|c|c|}
			\hline
			\textbf{Model} &
			\textbf{Subset} &
			\textbf{BLEU-2} &
			\textbf{ROUGE} &
			\textbf{CIDEr} &
			\textbf{BERT-F1} &
			\textbf{ACC} &
			\textbf{MRR} \\ \hline
			&
			\cellcolor[HTML]{C0C0C0}\textbf{First} &
			\cellcolor[HTML]{C0C0C0}20.80 (1.02) &
			\cellcolor[HTML]{C0C0C0}29.74 (1.59) &
			\cellcolor[HTML]{C0C0C0}41.26 (3.14) &
			\cellcolor[HTML]{C0C0C0}54.48 (1.38) &
			\cellcolor[HTML]{C0C0C0}57.12 (4.85) &
			\cellcolor[HTML]{C0C0C0}72.47 (3.19) \\ \cline{2-8}
			\multirow{-2}{*}{\textbf{Ref}} &
			\textbf{Later} &
			23.06 (1.20) &
			31.88 (1.66) &
			40.79 (2.83) &
			55.54 (1.40) &
			60.94 (2.67) &
			75.34 (1.59) \\ \hline
			&
			\cellcolor[HTML]{C0C0C0}\textbf{First} &
			\cellcolor[HTML]{C0C0C0}33.09 (0.79) &
			\cellcolor[HTML]{C0C0C0}42.32 (0.42) &
			\cellcolor[HTML]{C0C0C0}94.63 (2.12) &
			\cellcolor[HTML]{C0C0C0}62.55 (0.12) &
			\cellcolor[HTML]{C0C0C0}90.36 (1.73) &
			\cellcolor[HTML]{C0C0C0}94.49 (1.14) \\ \cline{2-8}
			\multirow{-2}{*}{\textbf{ReRef}} &
			\textbf{Later} &
			52.15 (1.19) &
			56.74 (0.63) &
			143.59 (5.84) &
			71.25 (0.39) &
			92.21 (0.73) &
			95.62 (0.45) \\ \cline{2-8}
			&
			\textbf{\it baseline} &
			36.66 (0.92) &
			45.37 (0.57) &
			96.41 (2.69) &
			64.13 (0.24) &
			90.14 (2.28) &
			94.38 (1.41) \\ \hline         
			&
			\cellcolor[HTML]{C0C0C0}\textbf{First} &
			\cellcolor[HTML]{C0C0C0}25.25 (0.40) &
			\cellcolor[HTML]{C0C0C0}33.31 (0.50) &
			\cellcolor[HTML]{C0C0C0}60.51 (1.21) &
			\cellcolor[HTML]{C0C0C0}57.61 (0.36) &
			\cellcolor[HTML]{C0C0C0}81.36 (0.53) &
			\cellcolor[HTML]{C0C0C0}88.70 (0.49) \\ \cline{2-8}
			\multirow{-2}{*}{\textbf{Copy}} &
			\textbf{Later} &
			43.08 (0.36) &
			48.79 (0.41) &
			128.45 (1.98) &
			66.07 (0.17) &
			83.96 (0.53) &
			90.60 (0.32) \\ \hline
		\end{tabular}%
	}
	\caption{Test set scores of the generation models (averaged over 5 runs) for first vs.~subsequent references, including word-overlap metrics, BERTScore F1, and accuracy/MRR obtained by our resolution model on the generated utterances. ReRef \emph{baseline} uses the first generated description verbatim in all later mentions.
All differences across model types are statistically significant ($p< 0.001$, independent samples $t$-test).}
	\label{tab:test_breakdown}
\end{table*}

%------------------------------------
\subsection{Generation Model Results}
%------------------------------------

As we did for the reference resolution model, we break down the test set into 
first referring utterances in a chain and subsequent references, for which generation is conditioned on 
a previous utterance.
The outcomes of this breakdown are provided in Table~\ref{tab:test_breakdown}, where we report the test set performances of our three generation models. Overall results on the validation set are available in Appendix \ref{sec:app_valres}.

{\bf ReRef} obtains the highest scores across all measures, followed by {\bf Copy}, while
{\bf Ref} achieves substantially lower results. 
Regarding the comparison between first and subsequent references,  the context-aware models {\bf ReRef} and {\bf Copy} attain significantly higher results when generating later mentions vs.~first descriptions ($p< 0.001$, independent samples $t$-test). As expected, no significant differences are observed in this respect for {\bf Ref}.\footnote{While first descriptions do not require linguistic context, {\bf ReRef} and {\bf Copy} perform better on first description generation than {\bf Ref}. This is likely due to their higher complexity.}

As for the communicative effectiveness of the generated utterances as measured by our resolution model, both accuracy and MRR are particularly high (over 90\%) for {\bf ReRef}. Across all model types, generated subsequent references are easier to resolve by the model, in line with the pattern observed in Table~\ref{tab:resolution} for the human data. 

All in all, the addition of the copy mechanism does not provide improvements over {\bf ReRef}'s performance that can be detected with the current evaluation measures. We do find, however, that the {\bf Copy} model uses a substantially larger vocabulary than {\bf ReRef}: 1,791 word types vs.~760 (the human vocabulary size on the test set is 2,332, while {\bf Ref} only uses 366 word types). An inspection of the vocabularies shows that {\bf Copy} 
does generate a good deal of low-frequency words, in line with what is expected from the dedicated copy mechanism (less desirably, this also includes words with spelling errors).
Further analysis also shows  that {\bf Copy} generates utterances that include more repetitions: 18\% of the  utterances generated by {\bf Copy} in the test set contain two identical content words e.g. \emph{``do you have the runway runway woman?''},
while only 7\% of those generated by {\bf ReRef} do.\footnote{The {\bf Ref} model is even more repetitive: 21\% of the generated utterances contain repeated content words.} 
Adding a means to control for repetitions, such as the `coverage' mechanism by~\citet{GTTP}, could be worth exploring in the future.

We compare our best performing model {\bf ReRef} to a baseline consisting in
reusing the first generated utterance verbatim in later mentions. In this case,
the model does not learn how to reuse previous referring utterances 
 taking into account the changing visual context, but simply 
keeps repeating the first description it has generated. We expect this baseline to be relatively strong
given that experiments in the lab have shown that dialogue participants may stick to an agreed description even when some properties are not strictly needed to distinguish the referent in a new visual context \cite{BrennanClark1996,brown2015people}.
The results (reported in Table~\ref{tab:test_breakdown} \emph{baseline}) show that the model significantly outperforms this baseline when generating later mentions.

Overall, our results confirm that referring utterances do evolve during a dialogue and indicate that the models that exploit the conversational context are able to learn some of the subtle modifications 
involved in the re-referring process. In the next section, we look into the linguistic patterns that characterise this process.

% !TEX root = main_pbgen.tex

%------ FIGURE--------

\begin{figure*}[t!]\centering
	\subfloat[Givenness markers]{{\includegraphics[height=3.6cm]{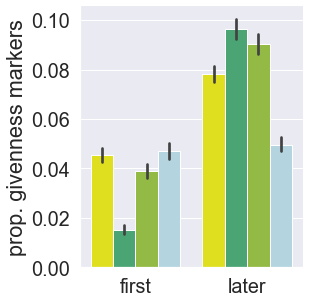} \label{fig:lingdef}}}%
	\qquad
	\subfloat[Proportion of nouns]{{\includegraphics[height=3.6cm]{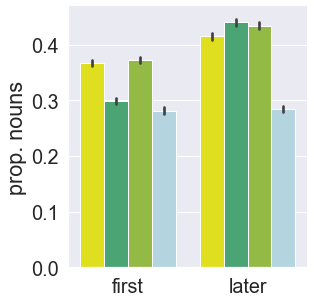} \label{fig:lingnouns}}}%
	\qquad
	\subfloat[Lexical entrainment in later references ]{{\includegraphics[height=3.6cm]{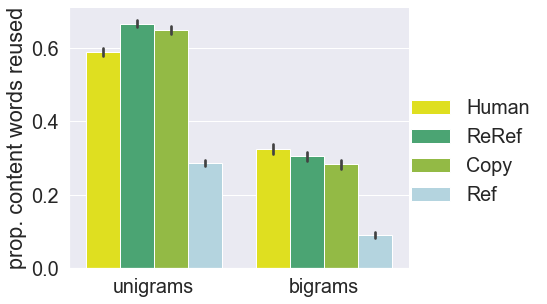} \label{fig:lingreuse}}}%
	\caption{Linguistic patterns in human referring utterances and in referring utterances generated by our three models. Givenness markers and proportion of nouns per utterance are displayed for first and later references.}
	\label{fig_linguistic_trends}
\end{figure*}

%------ END FIGURE--------

%==================================
\section{Linguistic Analysis}
\label{sec:analysis}
%==================================

We analyse the linguistic properties of the utterances generated by the best performing run of each of our 
models and compare them with patterns observed in the human data.
Extensive descriptive statistics are available in Appendix~\ref{sec:app_analysis}.

%---------------------------------------
\subsection{Main Trends}
%---------------------------------------
%---------------------------------------
\paragraph{Givenness markers}
%---------------------------------------
We first look into the use of markers of new vs.~given information, in particular indefinite and definite articles as well as particles such as \emph{again} or \emph{before} (as in \emph{``I have the X one again"} or \emph{``the X from before''}), which are anaphoric and presuppose that an image has been discussed previously in the dialogue.
Figure~\ref{fig:lingdef} shows the proportion of givenness markers (\emph{the, one, same, again, also, before}) in first vs.~subsequent references. Not surprisingly, this proportion increases in the
human subsequent references. {\bf ReRef} and {\bf Copy} both display an amplified version of this trend, while {\bf Ref}, which cannot capture any 
given information, shows no difference.

%---------------------------------------
\paragraph{Reduction}
%---------------------------------------
Regarding referring utterance length, we observe a significant shortening in subsequent mentions in human dialogues (11.3 vs.~8.3 tokens on average in first and subsequent mentions, respectively). This shortening is also observed in the utterances generated by {\bf ReRef} (11.3 vs.~7.2) and {\bf Copy} (10.8 vs.~7.8). {\bf Ref} tends to generate longer utterances across the board (13.7 vs.~13.6).

Shortening may be linked to compression, i.e., to an increase in information density~\cite{shannon1948mathematical}. To analyse this aspect, we consider the proportion of content words in the utterances, since such proportion can capture mechanisms such as syntactic reduction (e.g., the removal of the complementiser \textit{that}), which has been shown to be a good predictor of information density increase~\cite{jaeger2007speakers}.
\citet{haber2019photobook} reported a rise in the proportion of content words for all utterance types in later rounds of the PhotoBook games. We also observe such an increase in our referring utterance chains, and a similar trend is exhibited as well by the output of the {\bf ReRef} and {\bf Copy} models: In particular, generated subsequent references contain a significantly higher proportion of nouns and adjectives compared to first descriptions.
Figure~\ref{fig:lingnouns} shows this pattern for nouns, which are the most prominent type of content word in our data.

%---------------------------------------
\paragraph{Entrainment}
%---------------------------------------

In order to analyse the presence of lexical entrainment, we compute the proportion of expressions in subsequent references that are reused from the previous mention. We compare reuse at the level of unigrams and bigrams. Figure~\ref{fig:lingreuse} shows this information focusing on content words. Around 60\% of content tokens are reused by humans. The proportion is even higher in the utterances generated by our context-aware models.
Digging deeper into the types of content tokens being reused, we find that nouns are reused 
significantly more than other parts of speech by humans.
This is also the case in the subsequent references generated by the  {\bf ReRef} and {\bf Copy} models.

Humans also reuse a substantial proportion of content word bigrams---as do, to a smaller degree, the context-aware models.
For example, given the gold description \emph{``pink bowls rice and broccoli salad next to it''}, {\bf ReRef} generates the subsequent reference \emph{``pink bowls again''}.
Noun-noun compounds are a particularly interesting case of such bigrams, which we qualitatively analyse below.

%---------------------------------------
\subsection{A Case Study: Noun-Noun Compounds}
%---------------------------------------

A partial manual inspection of 
the human utterances in our chains reveals that, as they proceed in the dialogue, 
participants tend to produce referring expressions consisting of a noun-noun compound.\footnote{This is consistent with the fact that the proportion of noun-noun bigrams is significantly higher in subsequent references (0.05 vs.~0.08 on average in first and subsequent references, respectively;  $p < 0.001$ independent sample $t$-test).} 
For example, in Figure~\ref{fig:extraction} we observe the compound
\emph{``camera guy"} being uttered after the previous mention \emph{``\underline{guy} with \underline{camera}"}. (reused nouns are underlined).
Another example is \emph{``wine glass dog"} in Figure~\ref{fig:ex1}.
This is in line with~\citet{downing1977}, who argues that novel (i.e., not yet lexicalised) noun-noun compounds can be built by speakers on the fly based on a temporary, implicit relationship tying the two nouns, e.g., `the guy \emph{taking a picture with} a camera'. 
Such noun-noun compounds are thus prototypical examples 
of reuse and reduction: On the one hand, the novel interpretation~\cite[which needs to be pragmatically informative, diagnostic, and plausible;][]{costello2000efficient} can only arise from the established common ground between speakers;
on the other hand, compounds are naturally shorter than the `source' expression since they leave implicit the relation between the nouns.

\begin{figure}[t!]\centering
	\small
	\includegraphics[height=2.6cm]{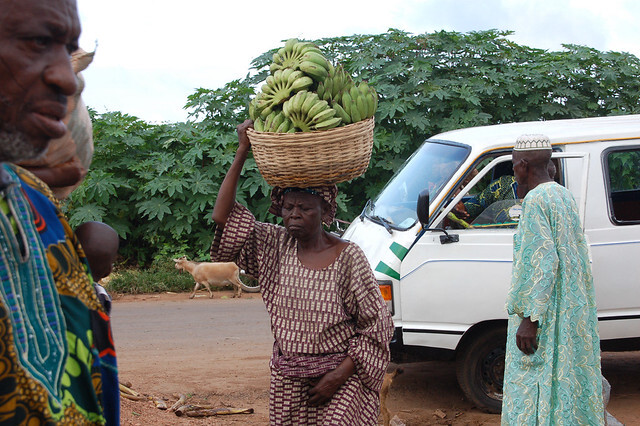}
	\includegraphics[height=2.6cm]{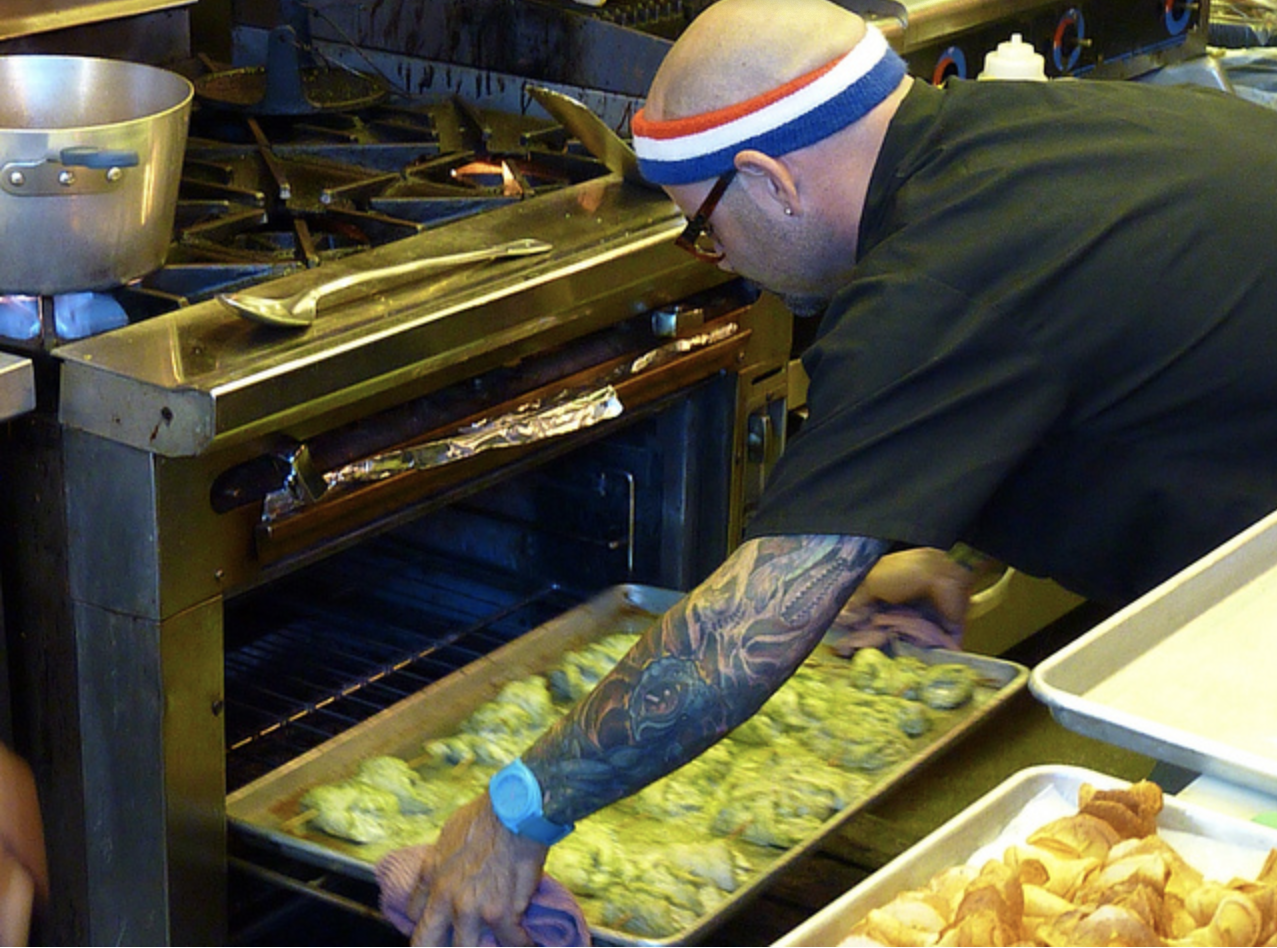} \\
	\vspace*{5pt}
	\resizebox{\columnwidth}{!}{
	\begin{tabular}{ll}
	\textbf{\emph{P:}} \underline{lady} with \underline{basket}? \textcolor{white}{aaaaaaaaaa} & \textbf{\emph{P:}} do you have headband \underline{guy}? \\[5pt]
	$\leadsto$ \textbf{ReRef:}  basket lady? & $\leadsto$ \textbf{ReRef:} tattoo guy?	\\ %\hline
	\end{tabular}
	}
	\caption{Two examples from the test set where \textbf{ReRef} generates a noun-noun compound based on the previous human mention (\textbf{\emph{P}}). Left: a genuine \emph{reuse} case; right: a \emph{non-reuse} case. Reused words are underlined.}
	\label{fig:compounds}
\end{figure}

We check whether our best performing generation models produce compounds
as humans do, i.e.,
by reusing nouns that were previously mentioned while compressing the sentence.
We perform the analysis with a qualitative focus, by manually inspecting a subset of the generated utterances.\footnote{The subset is obtained by applying simple heuristics to the set of generated utterances, such as length and PoS tags.} In Figure~\ref{fig:compounds}, we show two noun-noun compounds generated by \textbf{ReRef}
(similar cases were observed for \textbf{Copy}).
The example on the left is a noun-noun compound, \emph{``basket lady''}, that is consistent with the dialogue context: both nouns are indeed reused from the previous mention. In contrast, the compound on the right does not build on the conversational history; the noun \emph{``tattoo"} is not in the previous mention and never uttered within the reference chain (not reported),
and thus may be perceived as breaking a conceptual pact \cite{metzing2003conceptual}.
The compound is grounded in the image, but not in the conversational context.

% !TEX root = main_pbgen.tex

%==================================
\section{Conclusion}
\label{sec:conclusion}
%==================================

We have addressed the generation of descriptions that are (1) discriminative with respect to the visual context and (2) grounded in the linguistic common ground established in previous mentions. 
To our knowledge, this is the first attempt at tackling this problem at the level of surface realisation within a multimodal dialogue context. 

We proposed an encoder-decoder model that is able to generate both first mentions and subsequent references by encoding the dialogue context in a multimodal fashion 
and dynamically attending over it. 
We showed that our best performing model is able to produce better, more effective referring utterances than a variant that is solely grounded in the visual context. Our analysis revealed that the generated utterances 
exhibit linguistic properties that are similar to those observed in the human utterances regarding reuse of words and reduction. 
Generating subsequent references with such properties has the potential to enhance user adaptation and successful communication in dialogue systems.

Yet, in our approach we abstracted away from important interactive aspects such as the collaborative nature of referring in dialogue \cite{ClarkWilkes-Gibbs1986}, which was considered by \citet{shore-skantze-2018-using} for the task of reference resolution.  
In the present work, we simplified the interactive aspects of reference by extracting referring utterances from the PhotoBook dialogues and framing the problem as that of generating the next referring utterance given the previous mention. 
We believe that the resulting dataset of referring utterance chains can be a useful resource to analyse and model other dialogue phenomena, such as saliency or partner specificity, both on language alone or on the interaction of language and vision.

%================================
\section*{Acknowledgments}
This project has received funding from the European Research Council (ERC) under the European Union's Horizon 2020 research and innovation programme (grant agreement No.~819455).

%=== REFERENCES ================
\bibliography{pbgen} 
\bibliographystyle{acl_natbib}

%=== APPENDICES =================
\appendix
\section*{Appendices}

% !TEX root = main_pbgen.tex

%==================================
\section{Reference chain extraction}
\label{app:data}
%==================================

For our generation task, we extract reference chains of single referring utterances from the PhotoBook dataset \citep{haber2019photobook}. Given a dialogue and a target image, a reference chain is comprised of utterances---maximum one per round---that refer to the target image in that dialogue. Due to the size of the PhotoBook dataset (see Section \ref{sec:pb}), we perform this procedure automatically, with a three-step heuristic method described in the following sections. The chain extraction code is available at \url{https://dmg-photobook.github.io}.

\begin{table*}[] \centering
	\resizebox{2\columnwidth}{!}{
	\begin{tabular}{|l|c|c|c|c|c|c|c|}
	\hline
		& \textbf{Chains} & \textbf{Utterances}  & \textbf{Unique utterances} & \textbf{Target images}  & \textbf{Image domains}  & \textbf{Chain length}  & \textbf{Utterance length} \\
		\hline
		\textbf{Train}           & 11540              & 28933    & 27288                      & 360  & 30          & 2.51(0.85)    &  8.71(4.66)              \\ \hline
\textbf{Validation}   & 2503                & 6252        & 6009                  & 360     & 30      & 2.50 (0.85)    &  8.82 (4.67)            \\ \hline
\textbf{Test}            & 2482                & 6155       & 5876                    & 360    & 30       & 2.48 (0.86)    &   8.77(4.68)              \\ \hline
\textbf{Extracted-20} & 327                & 824           & 807              & 199   & 24         & 2.52 (0.85)    &     9.50 (4.75)       \\ \hline
\textbf{Gold-20}  & 327                & 756        & 740                  & 199      & 24     &  2.31 (0.94)   &   9.47 (4.77)       \\ \hline
	\end{tabular}
	}
	\caption{Descriptive statistics of all portions of the extracted dataset of reference utterance chains. Gold-20 is a set of 20 hand-annotated PhotoBook dialogues, with referent labels linking utterances to the target image they describe (see Section \ref{sec:extraction}) whereas Extracted-20 are the reference chains extracted from the same 20 dialogues, as if they were not annotated. Duplicate utterances are due to chance: PhotoBook participants have uttered them in different dialogues, potentially to describe the same target image. Image domains refers to the number of MS COCO image categories covered by a dataset portion; the 360 PhotoBook images come from a total of 30 domains.}
	\label{tab:splits-sm}
\end{table*}

\paragraph{Extracting dialogue segments}
The goal of segment extraction is to identify all utterances that may include a description of a given target image. To identify relevant segments, we leverage the participants' recorded actions, i.e. selecting an image as common or different \citep[more details on the available metadata in][]{haber2019photobook}.  When an image is selected by a participant as \textit{common} in a dialogue round, we extract all utterances up to that point in the round as candidate referring expressions. We collect referring expressions for a given image in a dialogue starting from the round when \textit{both} speakers observe it. The speakers are then more likely to have established a conceptual pact (see Section \ref{sec:intro}). 

\paragraph{Scoring referring utterances}
In this second step, we assign a score to each utterance in the extracted segments indicating how likely it is for that utterance to be a description of a given image. To produce these scores, we use as reference the MS COCO image captioning dataset \cite{lin2014coco} and the Visual Genome dataset of scene graphs \citep{krishna2017visual}. All 360 pictures in PhotoBook are taken from MS COCO, so we have access to at least 5 captions for each target image.
Instead, the Visual Genome dataset provides detailed scene graphs for 37\% of the PhotoBook images.

To measure the similarity of a candidate utterance to a reference MS COCO caption, we use the BERTScore \citep{bert-score}. We experiment with BERTScore Precision, Recall, F1, and select BERTScore F1.
As, in our dialogue setting, utterances often contain lexical material that is not part of a referring expression, we filter out stopwords from both the captions and the utterances. We use spaCy's stop-word list for English from which we remove numerals and prepositions that encode spatial information.\footnote{The English stop-word list is available at \url{https://github.com/explosion/spaCy/blob/master/spacy/lang/en/stop_words.py} and our edits at \url{https://dmg-photobook.github.io}.}.
Furthermore, to capture dyad-specific variation in referring language, we add the utterance with the highest BERTScore in a round to the reference set, and use it as an additional caption for the following rounds.

To take into account visual attributes and relationships, for each image we collect attribute tokens $T_A(i)$ (e.g. \textit{leafy}, \textit{tree} from \textit{leafy(tree)}) and relationship tokens $T_R(i)$ (e.g. \textit{man}, \textit{playing}, \textit{frisbee} from \textit{playing(man, frisbee)}) from the Visual Genome dataset of scene graphs. We only consider the intersection $T_{VG}(i) = T_A(i) \cap T_R(i)$ between the sets of attribute and relationship tokens to retain only the most relevant tokens. The set difference $T_{VG}(i*) \setminus \bigcup\nolimits_{i=1, i \neq i*}^{12}$ between the Visual Genome tokens of the target image and the tokens of the 11 distractors is then used as a reference set. To score an utterance, we compute its METEOR score \citep{banerjee2005meteor}  with respect to this reference set. 
For all images annotated in the Visual Genome dataset, the final utterance score is the sum of BERTScore and METEOR.\footnote{We implement BERTScore and use NLTK's code for METEOR (\url{https://www.nltk.org/api/nltk.translate.html}). We set METEOR's alignment penalty to 0 as our references are unordered collections of tokens.}

\paragraph{Selecting referring utterances}
The last step, utterance selection, produces reference chains consisting of single utterances---maximum one per round. As PhotoBook dialogues are made up of five rounds, reference chains will have a minimum length of 1 and a maximum possible length of 5.
First, given an extracted dialogue segment, we discard all utterances produced by speakers who do not have that image in their visual context. Then, for each target image in the corresponding dialogue round, we collect a ranked candidate list of $n$ top-scoring utterances. As an utterance can be selected as a candidate for multiple images in the same round, we discard a candidate \emph{(utterance, image)} pair if its score is lower than that of any other \emph{(utterance', image)} pair in the same round.
Finally, we pick the utterance with the highest score among the remaining candidates.
For some images, all of the $n$ top-scoring utterances are assigned to other images, and with higher scores. This causes a slight decrease in the number of utterances in the extracted dataset. We set $n=4$ to minimise the number of discarded utterances. Table \ref{tab:splits-sm} reports relevant statistics for the dataset splits of our extracted reference utterance chains.

\section{Data processing for models}
\label{sec:proc}
We further process the dataset of automatically extracted utterance chains. Every utterance is uniquely identified by the game ID, round number, message number and the ID of the image that they refer to. From these utterances and their contexts, we build the data we feed into our models.

While providing the 6 candidate images to the reference resolution models, we also keep track of the respective histories of candidates (the last utterance up to that time in the game). 

As the distribution of the 6 images and the positions of the target is not uniform 
for each target-context pair, this may constitute a bias in the reference resolution model. 
Therefore, to overcome this, we shuffle the images in the context for all splits at the beginning of each epoch. 
In the generation models, this shuffling is done once at the beginning of training for all splits.

\subsection{BERT representations}
\label{sec:bert}
Since utilising pre-trained BERT models and representations has proven to be beneficial to many NLP tasks~\cite{devlin-etal-2019-bert}, we also decided to use BERT to encode the linguistic input in the reference resolution models. For this purpose, we use the BERT-base-uncased model and the tokeniser as provided in the HuggingFace's Transformers library~\cite{Wolf2019HuggingFacesTS}. The utterances are first encoded into the correct format for BERT models.
Afterwards, they 
go through the BERT model to produce the hidden states that correspond to the representations of each of the input wordpieces. Finally, all utterances are fed into the reference resolution model in the form of a 
set of BERT representations.

We also experimented with using BERT-large-uncased model as well as 
extracting hidden states from multiple layers and aggregating them. Neither option provided further improvements on the results we obtained with the final hidden states from the BERT-base-uncased model. Hence, we opted to use the base model's outputs, where each hidden state is of size 768.

\subsection{Embeddings from scratch}
\label{sec:embeds}
For the generation models where we do not use BERT representations, we create a vocabulary of tokens from the training set 
with the help of TweetTokenizer from the NLTK library\footnote{\url{https://www.nltk.org/api/nltk.tokenize.html}}. We then map 
the words that occurred only once in the training split to '$<$unk$>$'. This results in a vocabulary of size 2816 (including $<$pad$>$, $<$unk$>$, $<$sos$>$, and $<$eos$>$). In addition to these special tokens, we also add $<$nohs$>$ to point out that there was \textbf{no} \textbf{h}i\textbf{s}tory (no previous utterance) for the target image at that point in the game. This token is utilised 
in the models that base their generation on the previous utterance. 
An input of  $<$nohs$>$ means that what the generation model is expected 
to produce 
is the very first utterance for that image in the game.

The tokens in all 3 splits are converted to indices using this final vocabulary. For the copy model, we need to keep track of what the actual form of an $<$unk$>$ token is. For this purpose, we build a full vocabulary from the whole dataset to have access to every word in all splits in their actual surface forms. This vocabulary is of size 5793 (including all 5 special tokens mentioned above).

Since we do not want the generation model to output the $<$nohs$>$ token, the search space of the decoder does not include this token. 
The Copy model needs to keep track of unknown tokens in the previous utterance and map the previous utterance using an extended vocabulary so that the decoder would be able to `copy' from the input itself, rather than only generating words from the reduced 
vocabulary. 
Mapped expected next utterance is used in calculating the loss. Actual inputs to the encoder and the decoder still contain unknown words, as we do not maintain special embeddings for the surface forms of each of the unknown tokens.

%===========================
\section{Model architectures}
\label{sec:models}
%===========================
Below are more details about our generation models and our reference resolution model.

\subsection{Generation models}
\label{sec:generation}

In these models, we apply teacher forcing during training; therefore, a token embedding at timestep $t$ is the embedding of the expected token from the ground-truth utterance. During validation, the models use the embedding of the word they generated in the previous timestep.

\subsubsection{ReRef model}
This model obtains the visual input as in the Ref model (consisting of the context and the target). However, instead of initialising the decoder as in the prior model, here, this visual representation initialises the encoder. The encoder receives as input a sentence that was previously used in the same game to refer to the target image (or simply $<$nohs$>$, if there was no history for the target image in the game at that point). The embeddings of this input go through dropout. 

We concatenate the last hidden states of the forward and backwards directions of the BiLSTM encoder. This concatenated vector is then projected to hidden dimensions and used to initialise the 
decoder. The input to the decoder during training is an embedding of the ground-truth utterance. 

For the attention mechanism, each hidden output of the encoder ${h_{enc}}^t$ (concatenation of forward and backward hidden states for timestep $t$) goes through a linear layer that projects it from double the size of hidden dimensions to the attention dimensions. In addition, the current hidden state of the decoder $ {h_{dec}}^c$ is projected from the hidden dimensions to the attention dimensions.

\begin{equation}
enc^t = W_e {h_{enc}}^t
\end{equation}
\begin{equation}
dec^c = W_d {h_{dec}}^c
\end{equation}
\begin{equation}
e_t = v_a(tanh(enc^t + dec^c))
\end{equation}
Attention weights are calculated based on the sum of $enc^t$ and $dec^c$, on which we apply $tanh$ non-linearity and a linear layer. Padded tokens are masked and softmax is applied over all remaining encoder timesteps $i$:
\begin{equation}
a_i = softmax(e_i)
\end{equation}
\begin{equation}
h^* = \sum_i a_ih_{enc}^i
\end{equation}

To predict the word that the decoder will generate, we concatenate the decoder's current hidden state ${h_{dec}}^c$  with the weighted average from the encoder, i.e. encoder context vector $h^*$. This concatenation is projected to the size of the vocabulary minus 1, as we do not want the model to predict the $<$nohs$>$ token. 

\subsubsection{Copy model}

The encoder part of this model is the same as that of the model explained in the previous subsection. However, this model uses various versions of the input and the decoder is altered to accommodate the copy mechanism.

First of all, we keep track of the unknown tokens in the input to provide the ability to predict them in the decoder phase. For this, we map the input utterance to temporary indices in a new extended vocabulary. 
This extended vocabulary contains the unknown words existing in the input utterance in their original forms appended to the end of the original vocabulary. Since we do not want $<$nohs$>$ to be predicted, we take additional precautions when it exists in the encoder input. The decoder input stays the same with unknown embeddings; nevertheless, the target utterance can include temporary indices assigned to unknown words encountered in the given input utterance, so that we can calculate the loss according to them as well.

The attention mechanism works in the same manner as in the previous model. However, we change what comes afterwards in line with the copy mechanism, where the attention for each word in the input utterance is added to their generation probabilities in the vocabulary. Here, we 
scatter the attention scores for the temporary indices of unknown words onto the distribution of the extended vocabulary, as well. For this reason, we maintain multiple versions of the input and output (mapped to the reduced vocabulary and mapped to the full vocabulary), as well as keeping track of the set of unknown words in the previous utterance and their temporary indices. Crucial here is the calculation of the generation probability $p_{gen}$, which requires the addition of several more linear layers that process the encoder context vector $h_t^*$, decoder input $x_t$, and the current decoder state $s_t$. As compared to the calculation of $p_{gen}$ by \citet{GTTP}, we altered the formula for this value by adding $tanh$ non-linearities: $p_{gen} = \sigma ($tanh$(w_{h^*}^Th_t^*) + $tanh$(w_{s}^Ts_t) + $tanh$(w_{x}^Tx_t))$.

\subsection{Reference resolution model}
\label{sec:refres}

In this model, BERT embeddings go through a dropout layer,
then a linear layer projecting the size to hidden dimensions. Finally, ReLU is applied~\cite{relu}.

All 6 images in the context are concatenated and the concatenation goes through dropout, a linear layer and ReLU 
to produce the final visual context vector. We then concatenate each of the BERT representations with the visual context vector to obtain multimodal token representations. This multimodal vector goes through a linear layer 
and ReLU, which finalises the multimodal input vectors. 
The model then determines the attention to be paid to each of the multimdal vectors as indicated below:
\begin{equation}
e_i = v_a(tanh(W_eh_i))
\end{equation}
\noindent$h_i$ is the multimodal output for each token, $W_e$ is a linear layer projecting from hidden dimensions to attention dimensions, $v_a$ is a linear layer that projects the output from the attention dimensions to a scalar. The model than masks the pad tokens before applying softmax over $e_i$ scores to obtain the attention weights  $a_i$:
\begin{equation}
a_i = softmax(e_i)
\end{equation}
\noindent The final multimodally-encoded 
utterance representation is then the weighted average of all $h_i$, given their attention weights $a_i$:
\begin{equation}
h_L = \sum_i a_ih_i
\end{equation}
\noindent Candidate images also separately go through dropout, a linear layer and ReLU. 
Finally, we normalise the outcomes for each image separately with L2 normalisation.

The history of each candidate image is determined by looking at their respective chains in the given game. Crucially, we only look at the chain items that were uttered before the current utterance we are trying to resolve. We take only the last utterance in the history, if such a history exists for a candidate image. In this case, we take the average of the BERT representations in the last utterance for that image. This average then goes through dropout, a linear layer and ReLU. 

The final history representation for a candidate image is added to this image's final visual representation to obtain its final candidate representation. Please note that not all images in the context necessarily have histories associated with them. Therefore, some candidate representations will be multimodal, whereas the others will remain in the visual domain, with no linguistic history being added.

To determine the target image, we take the dot product between the candidate representations and the multimodally-encoded utterance representation. The candidate with highest value is then predicted to be the referent of the input utterance.

\paragraph{Ablation:} As an ablation of the model described above, we train another type of model where the history is not added to the candidate images. Hence, the candidates are always represented only in the visual modality.

\paragraph{Baseline:} This model only uses one-hot vectors based on image IDs. These vectors go through the same operations as the image features go through in the models described above (dropout, linear layer, ReLU and normalization). At the end, instead of a dot-product, the outputs for the candidates are projected to scalar values and the model tries to predict the target via applying softmax directly over these scalars.
\begin{table}[]\centering \small
	\begin{tabular}{|l|c|c|}
		\hline
		\multicolumn{1}{|c|}{\textbf{Model}} & \textbf{Runtime} \\ \hline
		\textbf{Baseline}                     & 1 hour \\ \hline
		\textbf{Proposed}              &  5.5 hours  \\ \hline
		\textbf{Ablation}                     & 2.8 hours \\ \hline
	\end{tabular}%
	\caption{Resolution: approximate training runtimes.}
	\label{tab:listener_runtime}
\end{table}

\begin{table}[]\centering  \small
	\begin{tabular}{|l|c|c|}
		\hline
		\multicolumn{1}{|c|}{\textbf{Model}} & \textbf{Runtime} \\ \hline
		\textbf{Ref}                     &  6.5h\\ \hline
		\textbf{ReRef}                     & 7.5h \\ \hline
		\textbf{Copy}                     &  14h  \\ \hline
	\end{tabular}%
	\caption{Generation: approximate training runtimes.} 
	\label{tab:speaker_runtime}
\end{table}

\section{Evaluation metrics}
\label{app_evalmetrics}
For the evaluation of the reference resolution models, we use accuracy and mean reciprocal rank (MRR) implemented by us. Accuracy is a stricter measure as it is either 0 or 1 for a given instance. 

For the generation models, we use the $compute\_metrics$ function provided in the library at \url{https://github.com/Maluuba/nlg-eval} to obtain corpus-level BLEU, ROUGE, and CIDEr. 

We also report BERTScore~\cite{bert-score} for the generation models. To obtain this score, we use the library provided by the authors at \url{https://github.com/Tiiiger/bert_score} and import the $score$ function in our evaluation scripts. We use the BERT-uncased-model, we do not apply rescaling to baseline or importance weighting. The hash code for BERTScore that we used in evaluation is `bert-base-uncased\_L9\_no-idf\_version=0.3.2(hug\_trans=2.6.0)'. We obtain precision, recall and F1 variants of BERTScore.

\begin{table}[t]\centering  \small
	\begin{tabular}{|l|c|c|}
		\hline
		\multicolumn{1}{|c|}{\textbf{Model}} & \textbf{Parameters} \\ \hline
		\textbf{Baseline}                     &  182K \\ \hline
		\textbf{Proposed}                & 8.9M \\ \hline
		\textbf{Ablation}                     & 8.5M \\ \hline
	\end{tabular}%
	\caption{Resolution models: number of parameters.} 
	\label{tab:listener_paramno}
\end{table}

\begin{table}[t]\centering  \small
	\begin{tabular}{|l|c|c|}
		\hline
		\multicolumn{1}{|c|}{\textbf{Model}} & \textbf{Parameters} \\ \hline
		\textbf{Ref}                     &  16.1M\\ \hline
		\textbf{ReRef}                     &  24.9M \\ \hline
		\textbf{Copy}                     & 24.0M \\ \hline
	\end{tabular}%
	\caption{Generation models: number of parameters.} 
	\label{tab:speaker_paramnp}
\end{table}

\section{Model configurations and reproducibility}
\label{sec:config}

The models are implemented in Python 3.7.5\footnote{\url{https://www.python.org/downloads/release/python-375/}} and PyTorch 1.4.1\footnote{\url{https://pytorch.org/}}. In training our models, we use the Adam optimizer~\cite{Adam} to minimize the Cross Entropy Loss with sum reduction.\footnote{Copy model in fact uses the Negative Log-Likelihood Loss that receives log-softmax probabilities. This is equivalent to Cross Entropy Loss with logits. } 

We experimented with learning rate (0.001, 0.0001, 0.00001), dimensions for the embeddings (512, 1024), hidden and attention dimensions (512, 1024), batch size (16, 32) and dropout probability (0.0, 0.3, 0.5). 
We selected the best configurations per model type via manual tuning.

We train each model type with their selected configuration with 5 different random seeds setting the random behaviour of PyTorch and NumPy. We also turn off the cuDNN benchmark and also set cuDNN to deterministic. 

In all the models, the biases in linear layers were set to 0 and the weights were uniformly sampled from the range (-0.1, 0.1). In the models that learn embeddings from scratch, embedding weights were initialised uniformly in the range (-0.1, 0.1). The hidden and cell states of the LSTMs were initialised with task-related input at the first timestep.

\paragraph{Computing infrastructure:} The models were trained and evaluated on a computer cluster with Debian Linux OS. No parallelization was implemented, each model used a single GPU GeForce 1080Ti, 11GB GDDR5X, with NVIDIA driver version 418.56 and CUDA version 10.1.

\paragraph{Average runtimes:} Please see Table \ref{tab:listener_runtime}
and \ref{tab:speaker_runtime}. These durations indicate the total approximate runtime of training. The best models are reached in a shorter amount of time.

\paragraph{Number of parameters in each model:} Please see Table \ref{tab:listener_paramno} and Table \ref{tab:speaker_paramnp}.

\subsection{Configurations of the reference resolution models}
\begin{table*}[h!]
	\small
	\centering
	\resizebox{0.9\textwidth}{!}{%
		\begin{tabular}{|l|c|c|c|c|c|c|}
			\hline
			\multicolumn{1}{|c|}{\textbf{Model}} & \textbf{BLEU-2} & \textbf{ROUGE} & \textbf{CIDEr} & \textbf{BERT-F1} & \textbf{ACC} & \textbf{MRR} \\ \hline
			\textbf{Ref}      & 22.40 (1.22) & 31.29 (1.56) & 41.26 (3.18)  & 55.24 (1.38) & 59.69 (3.48) & 74.41 (2.21) \\ \hline
			\textbf{ReRef} & 45.41 (0.89) & 51.14 (0.42) & 127.08 (4.17) & 67.94 (0.23) & 91.70 (1.09) & 95.32 (0.70) \\ \hline
			\textbf{Copy}      & 36.44 (0.31) & 43.00 (0.35) & 104.27 (1.16) & 62.93 (0.21) & 83.28 (0.77) & 90.07 (0.49) \\ \hline
		\end{tabular}%
	}
	\caption{Average metric scores of the 3 generation models on the validation set. We report the average of 5 runs and standard deviations in parentheses. ACC is the reference resolution accuracy of the sentences generated by the generation models and MRR is their mean reciprocal rank as obtained through our best reference resolution model.}
	\label{tab:val_speaker}
\end{table*}

We select the reference resolution models based on their performance in accurately predicting the correct target among 6 images. We also report MRR, as it also provides further information in terms of the ranking of the correct image among the distractors.

After hyperparameter search, we decided on a batch size of 32, a learning rate of 0.0001, attention and hidden dimensions both set to 512, and a dropout probability of 0.5 for the proposed reference resolution model. We trained the ablation model with the same settings. 

\begin{table}[h]\centering 
	\resizebox{0.9\columnwidth}{!}{%
		\begin{tabular}{| l | r | r | c|}\hline
			\bf Subset & \multicolumn{1}{c|}{\bf ACC} & \multicolumn{1}{c|}{\bf MRR} & \multicolumn{1}{c|}{\bf Instances}\\ \hline
			\bf First  & 81.85 (0.45)  & 88.88 (0.29)  & 2503 \\\hline
			\bf Later  & 88.51 (0.19) & 93.33 (0.12) & 3749 \\\hline
			\bf Overall & 85.85 (0.10) & 91.55 (0.07)& 6252\\\hline
		\end{tabular}%
	}
	\caption{Validation set scores of the reference resolution model: averages of 5 runs with the best configuration, with the standard deviations in parentheses.}
	\label{tab:val_resolution}
\end{table}

\subsection{Configurations of the generation models}
Best-performing generation models for each model type were selected based on their performance with respect to the F1 component of BERTScore. We also performed hyperparameter search for beam width used in decoding, after which we decided to use a beam width of 3. The best-performing model for each model type outperformed the other models in its own category over all metrics.

As revealed by hyperparameter search, all reported generation models use 1024 dimensions for embeddings and 512 dimensions for hidden and attention layers. They all use a learning rate of 0.0001. Ref and Copy models use a batch size of 32 and the ReRef model, 16. Ref and ReRef models use a dropout probability of 0.3, whereas the Copy model yielded better results without dropout.

\section{Results on the validation set}
\label{sec:app_valres}
For each model we report in the main text, we also provide the validation set performances in Table \ref{tab:val_speaker} for the generation and Table \ref{tab:val_resolution} for the resolution models.

\section{Linguistic measures}
\label{sec:app_analysis}

\begin{table*}
  \small
\centering
\begin{tabular}{l |rrr|rrr|rrr|rrr}
   & \multicolumn{3}{c}{\textit{Human}}&\multicolumn{3}{c}{\textit{ReRef}}& \multicolumn{3}{c}{\textit{Copy}} & \multicolumn{3}{c}{\textit{Ref}}\\
  \hline
    & first & later & $d$ & first & later & $d$ & first & later & $d$ & first & later & $d$\\
  \hline
  \textit{\textbf{Givenness}} & & &  & & &  & & &  & & & \\

  givenness & 0.05 & 0.08 & -0.36* & 0.02 & 0.10 & -0.89* & 0.04 & 0.09 & -0.53* & 0.05 & 0.05 & -0.03\\
  definite & 0.03 & 0.05 & -0.27* & 0.01 & 0.08 & -0.85* & 0.03 & 0.06 & -0.48* & 0.04 & 0.05 & -0.04\\
  seen & 0.01 & 0.03 & -0.26* & 0.00 & 0.02 & -0.43* & 0.01 & 0.03 & -0.29* & 0.00 & 0.00 & 0.03\\
  indefinite & 0.07 & 0.02 & 0.77* & 0.15 & 0.01 & 1.88* & 0.10 & 0.01 & 1.14* & 0.15 & 0.15 & 0.03\\
  \textit{\textbf{Compression}} & & &  & & &  & & &  & & & \\
  length\_c & 11.29 & 8.28 & 0.63* & 11.32 & 7.22 & 1.15* & 10.77 & 7.79 & 0.65* & 13.66 & 13.59 & 0.00\\
  prop content & 0.53 & 0.57 & -0.20* & 0.41 & 0.54 & -0.70* & 0.50 & 0.58 & -0.39* & 0.40 & 0.39 & 0.01\\
  prop noun & 0.37 & 0.41 & -0.29* & 0.30 & 0.44 & -0.86* & 0.37 & 0.43 & -0.37* & 0.28 & 0.28 & -0.01\\
  prop adj & 0.09 & 0.10 & -0.02 & 0.06 & 0.07 & -0.14* & 0.08 & 0.09 & -0.10* & 0.08 & 0.08 & 0.04\\
  prop verb & 0.13 & 0.11 & 0.12* & 0.19 & 0.11 & 0.76* & 0.13 & 0.12 & 0.12* & 0.17 & 0.17 & 0.01\\
\end{tabular}
\caption{Trends in Subsequent mentions across humans, ReRef, Copy and Ref. The presence of * indicates significant differences between first and later means, with $p < 0.001$. $d$ shows effect size measured by Cohen's $d$.
}
\label{tab:ling_first_prev}
\end{table*}

\begin{table*}
  \small
\centering
\begin{tabular}{l r|rrr|rrr|rrr}
    \multicolumn{2}{c}{\textit{Human}}&\multicolumn{3}{c}{\textit{ReRef}}& \multicolumn{3}{c}{\textit{Copy}} & \multicolumn{3}{c}{\textit{Ref}}\\
  \hline
    & mean & mean & d & p & mean & d & p & mean & d & p\\
   \hline
 \multicolumn{2}{c|}{\textit{Lexical Entrainment:}} & & & & & & & & & \\
\textit{reuse prop within mention:} & & & & & & & & & & \\
--reuse\_c & 0.562 & 0.660 & -0.334 & *** & 0.612 & -0.168 & *** & 0.320 & 0.868 & ***\\
--reuse\_bigrams\_c & 0.325 & 0.304 & 0.050 & * & 0.283 & 0.103 & *** & 0.091 & 0.682 & ***\\
\textit{reuse prop within reused:} & & & & & & & & & & \\
--noun & 0.701 & 0.746 & -0.161 & *** & 0.716 & -0.050 & * & 0.740 & -0.124 & ***\\
--adj & 0.158 & 0.146 & 0.054 & * & 0.146 & 0.057 & * & 0.180 & -0.079 & **\\
--verb & 0.095 & 0.066 & 0.165 & *** & 0.097 & -0.011 & 0.653 & 0.063 & 0.172 & ***\\
--NN bigrams & 0.064 & 0.051 & 0.069 & ** & 0.056 & 0.043 & 0.064 & 0.013 & 0.328 & ***\\
\end{tabular}
\caption{Human comparison with ReRef, Copy and Ref for givenness markers and Compression. The presence of * indicates a significant difference between the human mean and that of the model. \textit{(***: p $<$ 0.001, **: p $<$ 0.005, *: p $<$ 0.01)}
}
\label{tab:ling_human_model}
\end{table*}

The linguistic measures used were chosen to quantitatively explore whether artefacts of the compression, reuse and grounding present in the human utterances, as well as other human-like linguistic patterns, can be seen in the generated utterances. We compare performance of the generation models with regards to the similarity of their generated sentences to human traits, namely a) whether there is a change in token use between first and last mention (Table~\ref{tab:ling_first_prev}) and b) whether this relative distance, or the values in the first mention differ significantly between human and model references (Table~\ref{tab:ling_human_model}).

In the case of givenness markers, we measure this as the proportion of tokens which correspond to definite (\textit{the}), indefinite (\textit{some, a, an}) and other markers of the existence of shared context (\textit{again, before, one, same, also}) which occur in the utterance. In the case of compression, we measure the lengths of the utterances in terms of tokens, and content tokens (tokens which are not in the stopword list from from nltk version 3.4.5~\cite{loper2002nltk}. We also measure the proportion of content words in an utterance which correspond to nouns, verbs and adjectives. Finally, for entrainment, examining only later utterances (not the first referent to an image), we measure firstly what proportion of the utterance in question consists of reused unigrams and bigrams from the previous utterance. We also measure within the reused tokens, the proportion of which is made up of nouns, adjectives and verbs, in order to discover their relative importance in terms of reuse. These measures can all be found in Tables~\ref{tab:ling_first_prev} and~\ref{tab:ling_human_model}. For these analyses we compared the generated output from the best seed for each model variant. These were seeds 1, 1, and 24 for the Ref, Copy and ReRef models respectively. We report both effect size ($d$) as measured by Cohen's d, and p-value (*$p < 0.05$, **$p < 0.005$, ***$p < 0.001$) for each comparison. We use the Scipy stats package (scipy version 1.3.3. ) \emph{ttest\_ind} to perform the independent t-test, and our own implementation to calculate Cohen's $d$ effect size.

Additionally to check general fluency, we evaluate the coherence and vocabulary use of the models in comparison to humans. We measure \textit{Type Token Ratio (TTR)}, the proportion of unique tokens in an utterance. This can capture ungrammatical repetition patterns in the generation, and, if following human trends, should increase in subsequent mentions. Although both models have significantly lower TTR than the human data, ReRef, unlike Copy, shows a significant increase in subsequent mentions, with much higher TTR than Copy, even though both models show similar average utterance length for later utterances \emph{(ReRef: 7.22, Copy: 7.79)}. In terms of vocabulary, for the generated outputs, ReRef has a much smaller \emph{(first: 492, later: 705)} vocabulary than Copy \emph{(first: 1098, later: 1469)}, although these are both much lower than Human vocabulary size \emph{(first: 1836, later: 1727)} and show an increase rather than a decrease in later mentions.

Overall, Tables~\ref{tab:ling_first_prev} and~\ref{tab:ling_human_model} show that both of our context-aware speaker models ReRef and Copy are able to generate referring utterances which make use of the dialogue history in a manner akin to humans with respect to multiple aspects of language style.

Comparing the context-aware models, ReRef shows a stronger degree of shortening than Copy, with very similar levels of bigram reuse to humans
while Copy shows more similar traits to humans in terms of proportion of markers and PoS tags (as revealed by smaller effect sizes). In general, both models are successful at generating human-like utterances as we measure them, however it seems that while Copy does generate utterances with the most similar proportional similarities to humans and exhibits similar proportions of unigram reuse, it does so at the expense of coherence. In terms of content bigram reuse, Copy seems to be less selective in what it repeats from previous referring utterances than ReRef, most likely due to the increased overall level of repetition in the generation. ReRef on the other hand shows amplified versions of the human trends, yet very similar content bigram and noun-noun bigram reuse proportion to humans, while maintaining low levels of same content word repetition as well as a high TTR, which indicates that coherence is also maintained.

%\label{sec:appendix}
%Appendices are material that can be read, and include lemmas, formulas, proofs, and tables that are not critical to the reading and understanding of the paper.
%Appendices should be \textbf{uploaded as supplementary material} when submitting the paper for review.
%Upon acceptance, the appendices come after the references, as shown here.

\end{document}